\RequirePackage{fix-cm}
\RequirePackage{amsmath}
\documentclass[twocolumn]{svjour3}          
\smartqed  
\usepackage{graphicx}
\usepackage[utf8]{inputenc}
\usepackage[T1]{fontenc}
\usepackage{hyperref}

\usepackage{tabularx}
\newcolumntype{C}{>{\centering\arraybackslash}X}

\usepackage{multirow}
\usepackage{amssymb}
\usepackage{textcomp}
\usepackage{booktabs}

\usepackage{tikz}
\usetikzlibrary{shapes}
\usetikzlibrary{positioning,arrows}
\usetikzlibrary{external}
\tikzset{
  >=stealth',
}
\tikzset{%
  do path picture/.style={%
    path picture={%
      \pgfpointdiff{\pgfpointanchor{path picture bounding box}{south west}}%
        {\pgfpointanchor{path picture bounding box}{north east}}%
      \pgfgetlastxy\x\y%
      \tikzset{x=\x/2,y=\y/2}%
      #1
    }
  },
  sin wave/.style={do path picture={
    \draw [line cap=round] (-3/4,0)
      sin (-3/8,1/2) cos (0,0) sin (3/8,-1/2) cos (3/4,0);
  }},
  cross/.style={do path picture={
    \draw [line cap=round] (-1,-1) -- (1,1) (-1,1) -- (1,-1);
  }},
  plus/.style={do path picture={
    \draw [line cap=round] (-3/4,0) -- (3/4,0) (0,-3/4) -- (0,3/4);
  }}
}

\journalname{IJCARS}
\begin{document}


\title{Retinal OCT disease classification with variational autoencoder regularization}

\titlerunning{Retinal OCT classification with autoencoder regularization}

\author{Max-Heinrich Laves \and
        Sontje Ihler \and
        Lüder A. Kahrs \and
        Tobias Ortmaier
}



\date{Received: date / Accepted: date}

\maketitle


\section{Purpose}

According to the World Health Organization, 285 million people worldwide live with visual impairment.
The number of people affected by blindness has increased substantially due to increasing life expectance.
80 \% of all causes are considered to be avoidable or curable by early diagnosis.
The most commonly used imaging technique for diagnosis in ophthalmology is optical coherence tomography (OCT).
However, analysis of retinal OCT requires trained ophthalmologists and time, making a comprehensive early diagnosis unlikely.

A recent study established a diagnostic tool based on convolutional neural networks (CNN), which was trained on a large database of retinal OCT images \cite{kermany2018}.
The performance of the tool in classifying retinal conditions was on par to that of trained medical experts.
However, the training of these networks is based on an enormous amount of labeled data, which is expensive and difficult to obtain.
Therefore, this paper describes a method based on variational autoencoder regularization that improves classification performance when using a limited amount of labeled data.

\section{Methods}

\begin{figure}
	\centering
	\includegraphics[width=0.75\columnwidth]{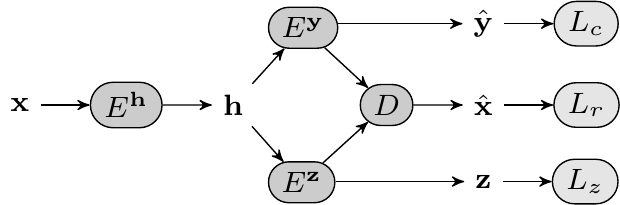} \\
	\vspace{1em}
	\includegraphics[width=0.95\columnwidth]{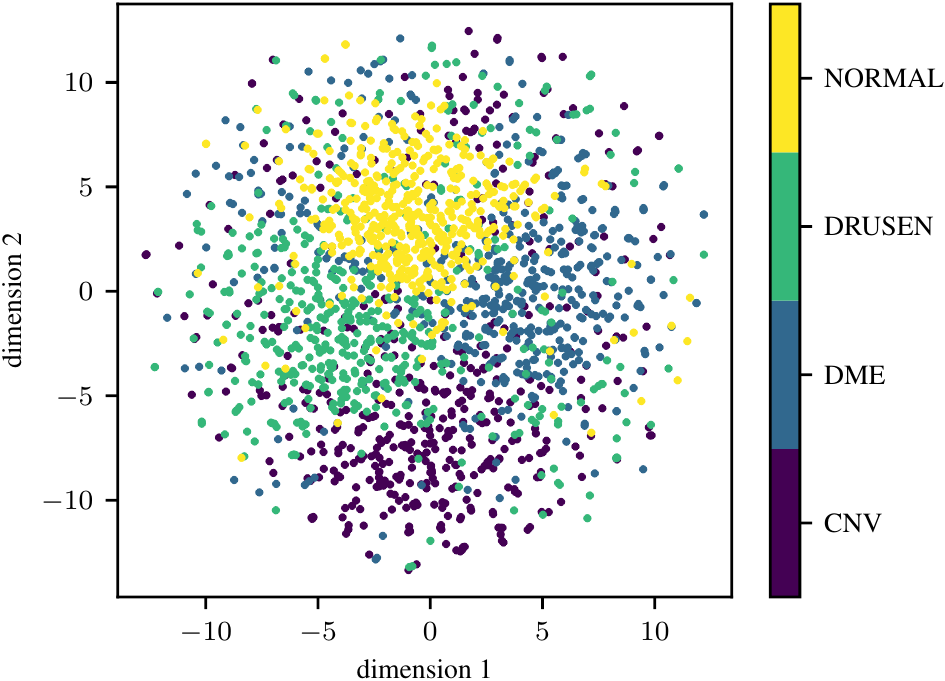}
	\caption{Top: The combined model consists of a VAE structure and a classification layer. The outputs $ \hat{\vec{y}}, \hat{\vec{x}}, \vec{z} $ are fed into different loss functions. Bottom: T-SNE projection of latent space from training set shows distinct clustering of the data.}
	\label{fig:1}
\end{figure}
The public dataset used in this study contains 84,484 retinal OCT images from 4,657 patients \cite{kermany2018}.
They are categorized in four classes showing different disease conditions (normal, drusen, DME and CNV).
The limited training set with employed ground truth labels contains 500 images per class and was randomly extracted from the dataset before training.
A validation set with 250 images per class was also randomly extracted.
The remaining 80,484 images were used for testing.
A simple preprocessing by automated center-cropping and resizing is performed.

This work uses a two-path CNN model combining a classification network with an autoencoder (AE) for regularization.
The key idea behind this is to prevent overfitting when using a limited training dataset size with small number of patients.
Basic AEs consist of two components. The encoder $ E^{\vec{z}} $ takes an input image $ \vec{x} $ and maps it from high dimension into low-dimensional, latent representation $ \vec{z} $.
This representation is then fed into the decoder $ D $ and mapped back to a reconstructed image $ \hat{\vec{x}} $.
The parameters of the combined encoder and decoder are optimized by minimizing a reconstruction error $ L_{r}(\hat{\vec{x}}, \vec{x}) $.
By regularization of the latent space, the performance of AEs can be improved.
In case of basic AEs, the distribution of $ \vec{z} $ is arbitrary.
To regularize this, variational autoencoders (VAEs) try to map all input images $ \vec{x} $ to a prior distribution.
First, additional encoders $ E^{\vec{\mu}} $ and $ E^{\vec{\sigma}} $ are added to output parameters $ \vec{\mu} $ and $ \vec{\sigma} $ of a parameterized posterior distribution.
Next, $ \vec{z} $ is created by drawing random samples by using the ``reparameterization trick'' $ \vec{z} = \vec{\mu} + \vec{\sigma} \vec{\varepsilon} $ with $ \vec{\varepsilon} \sim \mathcal{N}(\vec{0}, \vec{I}) $.
To bring the posterior distribution close to the prior, the Kullback-Leibler divergence (KLD) is added to the loss function.
In case of a standard normal distribution as prior, the KLD can be solved analytically \cite{kingma2013} as
\begin{equation}
    L_{\mathrm{z}}(\vec{\mu}, \vec{\sigma}^{2}) = - \frac{1}{2} \sum_{j=1}^{N} \left( 1 + \log(\sigma_{j}^{2}) - \mu_{j}^{2} - \sigma_{j}^{2} \right) ~ .
\end{equation}
For the final model (see Fig.~\ref{fig:1} (top)), a pretrained ResNet-34 architecture is used as $ E^{\vec{h}} $ to encode a feature vector $ \vec{h} \in \mathbb{R}^{1000} $, linear layers are used for $ E^{\vec{\mu}} $ and $ E^{\vec{\sigma}} $ with a latent size of $ N = 128 $, and the decoder of ErfNet is used for $ D $.
An additional linear layer $ E^{\vec{y}} $ is added to predict class labels from $ \vec{h} $.
The prediction $ \hat{\vec{y}} $ is also fed into the encoder by concatenation with $ \vec{h} $.
The combined model is fully trained by minimizing
\begin{equation}
  L = L_{c} + 0.1 \cdot L_{r} + 0.1 \cdot L_{\mathrm{z}} 
  \label{eq:combinedloss}
\end{equation}
using the Adam optimizer with a learning rate of $ \eta = 10^{-4} $ and a batch size of 64.
Pixel-wise mean squared error is used for $ L_{r} $ and cross entropy is used for $ L_{c} $.
Mean classification loss and accuracy are used as metrics to evaluate the trained model on the test set.

\section{Results}
\label{sec:results}

\begin{table}
	\centering
	\begin{tabular}{lcccc}
		\toprule
		  & precision & recall & F1 score\\
		 \cmidrule(lr){2-4}
		 ResNet-34             & 91.6 &    89.9 &    90.4 \\
		 ResNet-34+VAE (ours)  & \textbf{93.9} & \textbf{93.7} & \textbf{93.8} \\
		\bottomrule
	\end{tabular}
	\label{tab:results}
	\caption{Test set results of the VAE regularized approach compared to ResNet-34. Bold values denote best results.}
\end{table}
Results are summarized in Tab.~\ref{tab:results} and show superior classification performance compared to a pre-trained and fully fine-tuned baseline ResNet-34.
Fig.~\ref{fig:1} (bottom) shows the latent space reduced to two dimensions with t-dis\-tri\-bu\-ted stochastic neighbor embedding (t-SNE) for every sample of the training set.
Clustering in relation to the disease class is distinct.
Sampling the latent space can be used to generate new artificial OCT images.

\section{Conclusion}
\label{sec:conclusion}

Neural networks for disease classification on OCTs can benefit from regularization using variational autoencoders when trained with limited amount of patient data.
Especially in the medical imaging domain, data annotated by experts is expensive to obtain.
Further optimizations may be achieved by employing an additional denoising criterion or the use of different regularization techniques, such as adversarial autoencoders.

\section*{Disclosure of potential conflicts of Interest}

\paragraph{Conflict of Interest} The authors declare that they have no conflict of interest.

\paragraph{Funding} This research has received funding from the European Union as being part of the ERFE OPhonLas project.

\paragraph{Formal Consent} The medical images used in this article were made available to the public in a previous study \cite{kermany2018}, therefore formal consent is not required.

\bibliographystyle{spmpsci} 
\bibliography{literature}   

\end{document}